%% file: main.tex
\documentclass[conference]{IEEEtran}
\IEEEoverridecommandlockouts
\usepackage{cite}
\usepackage{amsmath,amssymb,amsfonts}
\usepackage{algorithmic}
\usepackage{graphicx}
\usepackage{textcomp}
\usepackage{xcolor}
\def\BibTeX{{\rm B\kern-.05em{\sc i\kern-.025em b}\kern-.08em
    T\kern-.1667em\lower.7ex\hbox{E}\kern-.125emX}}

\input{packages}

\begin{document}

% \title{Conference Paper Title*\\
\title{LLM-Based Automated Grading with Human-in-the-Loop}

\author{\IEEEauthorblockN{Yucheng Chu\IEEEauthorrefmark{1}}
\IEEEauthorblockA{\textit{Computer Science and Engineering} \\
\textit{Michigan State University}\\
East Lansing, USA \\
chuyuch2@msu.edu}
\and
\IEEEauthorblockN{Hang Li\IEEEauthorrefmark{1}}
\IEEEauthorblockA{\textit{Computer Science and Engineering} \\
\textit{Michigan State University}\\
East Lansing, USA \\
lihang4@msu.edu}
\and
\IEEEauthorblockN{Kaiqi Yang}
\IEEEauthorblockA{\textit{Computer Science and Engineering} \\
\textit{Michigan State University}\\
East Lansing, USA \\
kqyang@msu.edu}
\and
\IEEEauthorblockN{Yasemin Copur-Gencturk}
\IEEEauthorblockA{\textit{Rossier School of Education} \\
\textit{University of Southern California}\\
Los Angeles, USA \\
copurgen@usc.edu}
\and
\IEEEauthorblockN{Jiliang Tang}
\IEEEauthorblockA{\textit{Computer Science and Engineering} \\
\textit{Michigan State University}\\
East Lansing, USA \\
tangjili@msu.edu}
\thanks{\IEEEauthorrefmark{1}Both authors contribute equally.}

}

\maketitle

\begin{abstract}
The rise of artificial intelligence (AI) technologies, particularly large language models (LLMs), has brought significant advancements to the education field. Among various applications, automatic short answer grading (ASAG), which focuses on evaluating open-ended textual responses, has seen remarkable progress with LLMs. These models not only enhance grading performance compared to traditional ASAG approaches but also move beyond simple comparisons with predefined answers, enabling more sophisticated grading scenarios, such as rubric-based evaluation. However, existing LLM-powered methods still face challenges in achieving human-level grading performance in rubric-based assessments due to their reliance on fully automated approaches. In this work, we explore the potential of LLMs in ASAG tasks by leveraging their interactive capabilities through a human-in-the-loop (HITL) approach. Our proposed framework, GradeHITL, utilizes the generative properties of LLMs to pose questions to human experts, incorporating their insights to dynamically refine grading rubrics. This adaptive process significantly improves grading accuracy, outperforming existing methods and bringing ASAG closer to human-level evaluation.
\end{abstract}

\begin{IEEEkeywords}
Automatized Grading, Large Language Model, Human-in-the-Loop.
\end{IEEEkeywords}

\input{1intro}

\input{2related_work}

\input{3problem}

\input{4method}

\input{5experiment}

\input{6discussion}

\input{7conclusion}

% \section*{Acknowledgment}

% The preferred spelling of the word ``acknowledgment'' in America is without 
% an ``e'' after the ``g''. Avoid the stilted expression ``one of us (R. B. 
% G.) thanks $\ldots$''. Instead, try ``R. B. G. thanks$\ldots$''. Put sponsor 
% acknowledgments in the unnumbered footnote on the first page.

\bibliographystyle{IEEEtran}
\bibliography{a_ref}

\end{document}

%% file: packages.tex
\usepackage{graphicx} % Required for inserting images
\usepackage{hyperref}       % hyperlinks
\usepackage{url}            % simple URL typesetting

\usepackage{amsfonts}       % blackboard math symbols
\usepackage{bbm}
\usepackage{cite}
\usepackage{xcolor}         % colors
\usepackage{amsmath}
\usepackage{ulem}
\usepackage{multirow}
\usepackage{booktabs}
\usepackage{array,float}
\usepackage{threeparttable}
\usepackage{wrapfig}
\usepackage{quoting}
\usepackage{graphicx}
\usepackage{calc} 
\usepackage{enumitem}
\usepackage{colortbl}
\usepackage{bbding}

\usepackage{ulem}
\usepackage[most]{tcolorbox}
\tcbset{
    mybox/.style={
        colback=gray!10,    % Background color
        colframe=gray!50,   % Frame color
        fonttitle=\scriptsize\bfseries, % SMALL bold Title font
        boxrule=0.5pt,      % Thinner border
        title={#1},         % Title parameter
        left=1mm, right=1mm, top=0.5mm, bottom=0.5mm, % less padding
        boxsep=0.6mm,       % less internal padding
        enhanced,           % better rendering
        sharp corners=south % optional: squared bottom corners
    }
}

\usepackage{multirow}
\usepackage{makecell} 
\usepackage[linesnumbered,ruled,vlined]{algorithm2e}
\SetKwBlock{DoParallel}{do $l \gets 1$ \KwTo $L$ in parallel}{end}
\usepackage{hyperref}
\usepackage{soul}
\usepackage{tabularx}
\usepackage{longtable}
\usepackage{tikz}
\usepackage{xcolor}
\usepackage{calc}
\usepackage{setspace}
\usepackage{graphicx}
\usepackage{booktabs}
\definecolor{keywords}{RGB}{0,0,255}
\definecolor{comments}{RGB}{0,128,0}
\definecolor{strings}{RGB}{255,0,0}
\usepackage{listings}
\usepackage{bbm}

%% file: 1intro.tex
\section{Introduction}
\label{sec:introduction}
The recent advance of artificial intelligence (AI) technologies, such as large language models (LLMs), is revolutionizing various real-world application domains~\cite{rane2023contribution,zhao2024revolutionizing}. In the field of education, the application of LLMs offers significant benefits across various directions, including adaptive learning~\cite{li2024bringing}, teaching \& learning assistance~\cite{wang2024large}. Among these applications, automatic short answer grading (ASAG), which focuses on evaluating open-ended textual answers, has made remarkable progress with the advent of LLMs. LLMs not only enhance grading performance compared to traditional ASAG approaches but also go beyond simple comparisons with predefined answers, enabling more complex grading scenarios, such as rubric-based grading~\cite{senanayake2024rubric}. Taking advantages of its exceptional capabilities in logical reasoning, language understanding, and prior knowledge, LLM have been extensively explored in recent studies for various rubric grading scenarios, achieving promising results~\cite{chu2024llm,schneider2023towards,chu2025enhancing}. While these studies highlight the effectiveness of LLMs for ASAG, several limitations persist due to their reliance on fully automated approaches. For instance, rubric texts often contain jargon or domain-specific terms that lack clear explanations. Fully automated methods struggle to accurately interpret these terms based solely on labeled examples, which leads to performance bottlenecks in the final grading outcomes. Additionally, the inherent complexity of language expression introduces challenges in achieving robust and controllable ASAG systems. Small variations in input texts can lead to significant changes in output results, further complicating the optimization and execution of ASAG in a fully automated manner. These limitations underscore the need for methods that integrate human feedback and domain knowledge to address the shortcomings of purely automated approaches.

To address these issues, in this paper, we explore the potential of LLMs for ASAG tasks by leveraging their interactive features through a human-in-the-loop (HITL)~\cite{wu2022survey} approach. Our proposed LLM-powered ASAG framework incorporates HITL design, enabling LLMs not only to passively output final grades but also to actively raise questions about rubrics or their grading errors. By incorporating answers from human experts, the framework adaptively optimizes grading rubrics, leading to significant improvements in grading performance compared to existing methods. Moreover, the interaction between humans and LLMs ensures highly controllable grading standards, a critical requirement for applications in the education field~\cite{yan2024practical}. However, questions generated by LLMs during this process are not always of high quality~\cite{zhang2025can}. To address this challenge, we further introduce a reinforcement learning (RL)-based HITL Q\&A selection method that filters out low-quality questions. Using the proximal policy optimization (PPO) method~\cite{sutton2018reinforcement}, our model is trained to identify and prioritize valuable questions, thereby reducing noise during the HITL process and enhancing overall performance. Finally, to demonstrate the effectiveness of our framework, we implement it alongside a state-of-the-art LLM-powered rubric grading method, GradeOpt~\cite{chu2024llm}. This method uses automatic prompt optimization to address challenging grading tasks involving questions designed to assess the knowledge and skills required for mathematics teaching~\cite{copur2022mathematics}. Through comprehensive experiments on 6 selected questions, we show that our proposed framework outperforms existing methods by effectively leveraging the interactive features of LLMs and HITL mechanism.

%% file: 2related_work.tex
\section{Related Work}

\subsection{Automatic Short-Answer Grading}
\vspace{-1mm}
Automatic short-answer grading (ASAG) has become a key research area for reducing teachers' workloads. Early ASAG methods relied on hand-crafted textual features and pattern-matching algorithms~\cite{burrows2015eras}. With the advent of deep learning~\cite{lecun2015deep} and advances in natural language processing (NLP) such as transformers~\cite{vaswani2017attention}, recent ASAG research has shifted toward semantic-based grading methods~\cite{mohler2009text,camus2020investigating,bonthu2021automated}. The emergence of pre-trained language models (PLMs), such as BERT~\cite{devlin2018bert} and RoBERTa~\cite{liu2019roberta}, further advanced ASAG by eliminating the need for traditional training paradigms that required large labeled datasets. Leveraging fine-tuning techniques, PLM-based approaches have expanded ASAG to encompass languages beyond English~\cite{sawatzki2021deep,chang2024automatic} and questions beyond traditional reading comprehension tasks~\cite{lertchaturaporn2024automated,messer2024automated}. More recently, LLMs have revolutionized ASAG by addressing key challenges, including rubric-based grading—a limitation of earlier comparison-based methods~\cite{senanayake2024rubric,chu2024llm,chu2025enhancing}. LLMs demonstrate human-like reasoning, deep language understanding, and generate actionable feedback. Crucially, they not only assess the answer content but also interpret and validate the students' underlying problem-solving ideas~\cite{chu2024llm}.

\subsection{Human-in-the-Loop with LLMs}
\vspace{-1mm}
Human-in-the-Loop (HITL) is a technique that incorporates human oversight, intervention, or feedback into AI systems to enhance their performance, ensure reliability, and address ethical concerns~\cite{wu2022survey}. LLMs, as some of the most advanced AI algorithms, have been the focus of recent exploration of the integration of HITL. For instance, Reinforcement Learning from Human Feedback (RLHF)~\cite{ouyang2022training} is a successful implementation of HITL with LLMs. By collecting human preferences for various candidate responses generated by LLMs, RLHF algorithms~\cite{christiano2017deep,rafailov2024direct} leverage RL to align the generative behavior of LLMs with human interests, significantly increasing the acceptance rate of their outputs. Beyond using human input to generate fine-tuning data, recent studies have also investigated the role of human supervision during the inference process. For example, Xiao et al., employed HITL to support path-planning tasks in LLMs, proposing LLM A*~\cite{xiao2023llm}. This approach achieved few-shot, near-optimal path planning compared to data-driven models such as PPO. Similarly, Cai et al., explored the combination of HITL with chain-of-thought (CoT) reasoning~\cite{wei2022chain} by introducing the Manual Correction System (MCS), which identifies when and how manual corrections of rationales can effectively enhance LLMs’ reasoning capabilities~\cite{cai2023human}. Lastly, Yang et al. collected human annotator feedback on LLM-based machine translation results~\cite{yang2023human}. They utilized LLMs' in-context learning capabilities to retrieve relevant human feedback, enabling the models to refine and improve translation quality in subsequent iterations.

%% file: 3problem.tex
\section{Problem Statement}
\label{sec:problem}
% \vspace{-2mm}

The output of ASAG ($\hat{y}$) generally falls into two categories: numerical or categorical scores. For simplicity, we formulate the ASAG task as a text classification problem, where a given short answer is assigned to one of the discrete score categories $\{c_i \mid i = 1, \dots, C\}$, with $C$ representing the total number of categories. The function $\mathcal{F}$, which defines the ASAG system, varies in form depending on the backbone model employed. In this study, we implement grading using LLMs, where $\mathcal{F}$ is expressed as $\mathcal{F}_{\theta}(G \| a)$. Here, $G$ denotes the grading rubric, $\theta$ represents the LLM parameters, and $\|$ signifies text concatenation. Instead of fine-tuning $\theta$, which is computationally expensive and requires large-scale labeled datasets, we focus on optimizing $G$ to improve grading performance. Building on prior work~\cite{chu2024llm}, we leverage the reflective and self-refining capabilities of LLMs to generate an optimized grading rubric $G^*$ that enhances model accuracy. However, unlike GradeOpt, our approach incorporates human feedback $H$ into the optimization process, aiming to improve both grading performance and rubric controllability.

%% file: 4method.tex
\section{Method}
% \vspace{-2mm}

\begin{figure*}[!btph]
    \centering
    \includegraphics[width=0.55\linewidth]{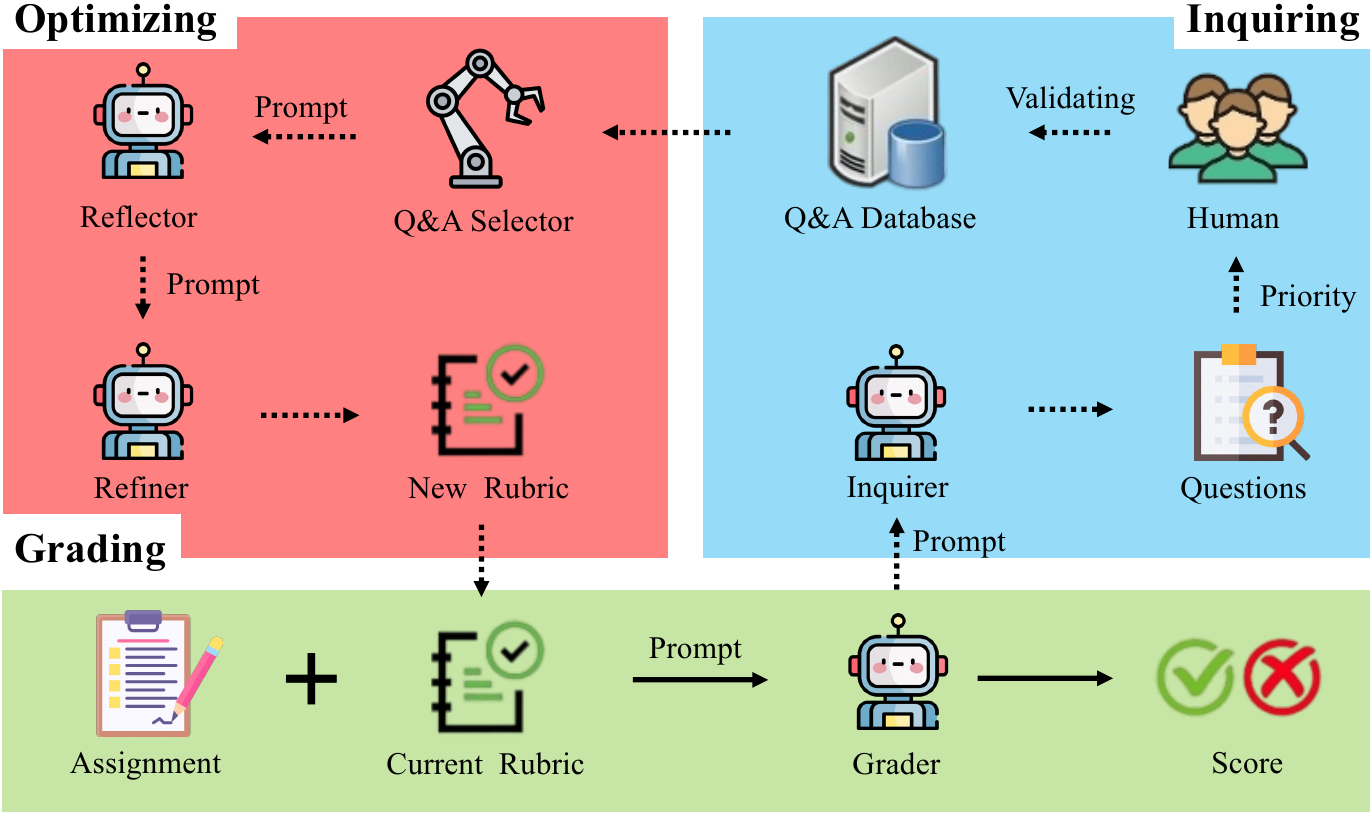}
    \vspace{-2mm}
    \caption{Illustration of GradeHITL}
    \label{fig:framework_rag}
\end{figure*}

In this section, we introduce GradeHITL, a human-in-the-loop ASAG framework that builds upon the rubric optimization approach of the prior LLM-powered ASAG framework, GradeOpt~\cite{chu2024llm}. While GradeOpt focuses on fully automated rubric optimization, GradeHITL incorporates human feedback to overcome the limitations of automation, improving grading accuracy and reliability. The framework consists of three key components: Grading, Inquiring, and Optimizing. An overview of our framework is illustrated in Figure~\ref{fig:framework_rag}. The Grading component serves as the system’s core, where LLMs process rubrics and responses to generate grading results, denoted as $\hat{y}$. These results can be used directly or compared with expert-annotated grades to inform the subsequent Inquiring and Optimizing stages. In the Inquiring stage, the LLM inquirer leverages its generative capabilities to pose targeted questions about the rubric text $G$. To minimize human workload, only the most relevant and impactful questions are selected and presented to human experts, whose responses clarify ambiguities and refine the rubric. Finally, in the Optimizing stage, the human-computer Q\&A collected during inquiry is integrated into an iterative rubric refinement process. By incorporating expert feedback, the system systematically identifies deficiencies and enhances the rubric’s effectiveness for LLM-based grading. The optimized rubric is then reintegrated into the system for continuous improvement. 
% \vspace{-2mm}

\subsection{Grading}
\vspace{-1mm}

The \textit{Grading} component maps the input response $ a $ to the score $ c_i $ based on the given rubric $ G $. As introduced in Section~\ref{sec:problem}, LLM-powered graders typically concatenate the rubric and response text, then directly output the grading results. To mitigate simple LLM errors and fully leverage their potential, we incorporate CoT prompting~\cite{wei2023cot}. This approach encourages LLMs to provide both judgments and intermediate reasoning steps, making their grading process more transparent and aligned with human evaluation. The intermediate reasoning not only justifies grading decisions but also highlights potential flaws in the rubric, aiding its refinement. The prompt for the LLM-powered grader is shown in Figure~\ref{fig:grade_prompt}. During testing, raw outputs are processed using regular expressions to extract the final categorized grade. In the optimizing stage, the LLM grading results ($\hat{y_i}$) are compared with expert grading scores ($y_i$), and mismatched answers ($A_{error}=\{a_i | \mathcal{F}(G,a_i) \neq y_i \}$ error samples) are collected and forwarded to the optimizing component for further analysis. Additionally, the detailed reasoning chains and likelihood scores of each sample are passed along with the grading results to guide rubric improvements, ensuring more reliable and consistent grading outcomes.

\input{examples/grade_prompt}

\subsection{Inquiring}
\vspace{-1mm}

As mentioned in Section~\ref{sec:introduction}, one of the weaknesses of existing LLM-powered rubric grading methods is that rubric texts often contain jargon or domain-specific terms without clear explanations. This lack of clarity makes it challenging for LLMs to fully understand these terms, even with providing labeled samples. To address this issue, we propose a solution within our framework: the \textit{Inquirer}. This component actively generates questions about the rubric, helping LLM-powered graders identify and express areas of confusion. By involving human experts to provide answers, our goal is to offer additional support to the LLM grader, improving their understanding of the rubric and ultimately enhancing the accuracy of the grading results. Overall, the inquirer operates in two sequential steps: \textit{inquiry} and \textit{validation}. In the inquiry step, we instruct LLMs to generate questions based on the rubric. To implement this, we input the grading rubric $G$ alongside labeled demonstration examples (e.g., responses $a$ with labels $y$) and prompt LLMs to express any uncertainties regarding the rubric or the demonstrations. The inquirer is guided to raise questions on utilizing $G$ to grade demonstrations, which can be categorized into three main directions: seeking rubric clarification, inquiring about meaning of demonstration detail, and pinpointing the level of fulfillment to the expected concept. To reduce the human effort required to answer a large volume of questions, we develop a question-ranking method based on the LLM's grading confidence. This method prioritizes questions that raised from the grading results ($\hat{y}$) where LLM-powered graders which are less confident about, as these are likely to be more critical for improving the accuracy of the grading. Low-confidence grading results typically indicate areas where the LLM is unsure of how to grade a response properly, so these should take higher priority. Questions with higher priority are retained and passed to human graders for answers. In empirical studies, considering the extensive amount of questions, we adopt a LLM to act as the \textit{Answerer} who provides answers to the remaining questions. However, due to inherent randomness in the generation of questions and the diversity of textual expressions used by different individuals or AI agents, the effectiveness of each question-and-answer pair in improving grading performance is not guaranteed. If not handled appropriately, low-quality Q\&As may overwhelm key information, potentially undermining grading accuracy. 
To address this issue, we introduce the validation step, in which we assess the effectiveness of each Q\&A pair. The key indicator of effectiveness is the correctness of grading results after incorporating Q\&As into grading process. For each Q\&A pair, we use a validation dataset, concatenate the Q\&A with rubric and response, and observe the resulting grading accuracy. The validation prompt is the same as grading prompt as in Figure~\ref{fig:grade_prompt} since they serve the same purpose. We assume Q\&A pairs that lead to correct grading are valid, while those that result in incorrect grading are invalid. Based on this assumption, we filter out ineffective Q\&As. Finally, by collecting validated Q\&A pairs, we build an external database that can be used to optimize future grading processes.

\input{examples/question_prompt}

\subsection{Optimizing}
\vspace{-1mm}

The \textit{Optimizing} component refines the grading rubric to improve accuracy and consistency. Following the GradeOpt framework~\cite{chu2024llm}, it leverages the self-reflective capabilities of LLMs within a multi-agent system. The process begins by identifying responses with grading errors, allowing the LLM to analyze patterns, diagnose underlying issues, and propose refinements. To enhance this process, we integrate human-computer Q\&A interactions, enabling the LLM to clarify ambiguities and refine the rubric more effectively. This optimization is carried out through a multi-agent framework comprising three key agents, the Retriever, Reflector, and Refiner. These agents work in a pipeline, iteratively updating the rubric to improve grading reliability and consistency.

% \vspace{-1mm}
\begin{figure*}[!btph]
    \centering
    \includegraphics[width=.55\linewidth]{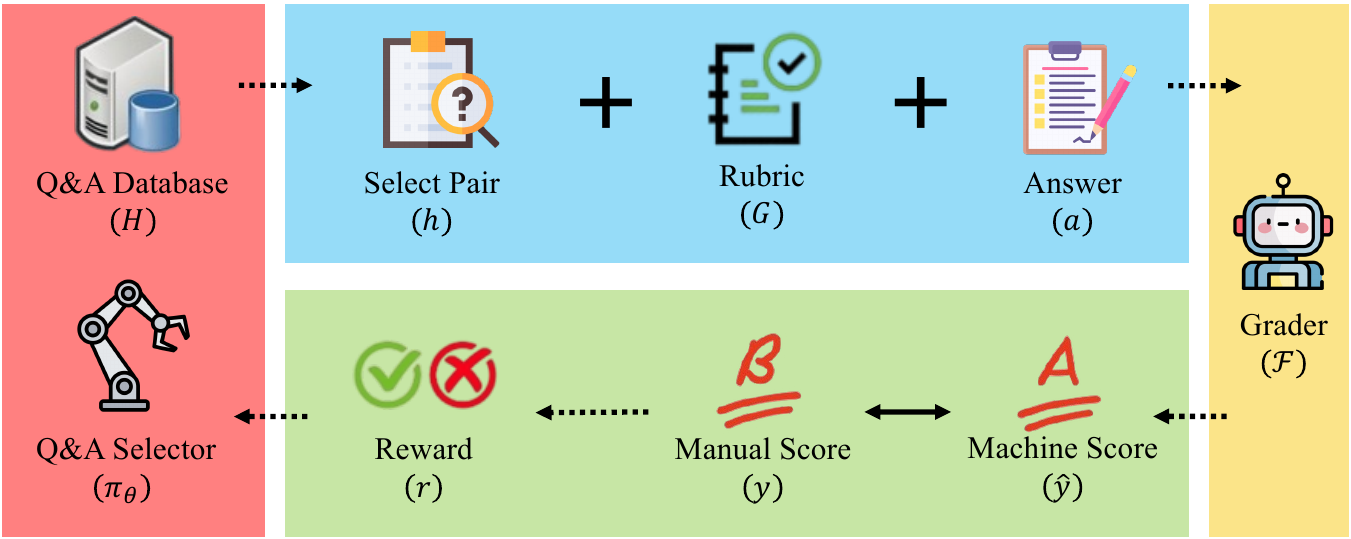}
    \caption{Illustration of reinforcement learning based Q\&A selector.}
    \label{fig:rl_diag}
\end{figure*}
% \vspace{-2mm}

\subsubsection{Retriever}

The role of the \textit{Retriever} is to retrieve valuable Q\&A pairs from a Q\&A database $H$ to provide human interpretations that assist in the reflection step. Since the context window of LLMs is limited, it is infeasible to incorporate all Q\&A pairs into the model. Moreover, only a few samples are extracted for reflection in each iteration, and irrelevant Q\&A pairs may distract the \textit{Reflector}. To address this, we propose to develop a retriever to find the useful Q\&A pairs from the dataset. There are a few heuristic ways to achieve the goal. For example, it is viable to leverage embedding models (e.g., Sentence-BERT~\cite{reimers2019sentencebert}) to find the most relevant Q\&A pairs based on semantic similarity. However, in practice, these manual heuristic method is not guaranteed to help the optimization to its best result. To solve that issue, we get inspiration from the existing study on in-context learning~\cite{lu2022dynamic} and propose to use RL for selecting the top-k pairs with the highest effectiveness in helping grader to achieve better grading accuracy. To be specific, we train a policy model that selects the top-k Q\&A from the candidate pool. The selected Q\&A record $h$ is added to \textit{Reflector}'s prompt, aiding in the generation of error feedbacks. \textit{Refiner} generates a better guideline by addressing the errors and confusions. The reward of $h$ is computed by evaluating \textit{Grader}'s correctness with the refined guideline on the mistaken samples. That is, the refined guideline that yields a correct prediction $\hat{y}$ awards $r$ with +1, while the still incorrect prediction $\hat{y}$ leads to a punishment of -1. The policy is updated according to the rewards of each batch to augment the probability of helpful pairs being selected and reduce those that may harm performance. The selecting model and the reward function can be formulated as follows: 

\vspace{-5mm}

{\scriptsize
\begin{align}
    \nonumber
    &e^{(x)} =\mathcal{E}(x) =\mathcal{E}(q^{(x)}||a^{(x)}),\ e^{(h)} =\mathcal{E}(h)=\mathcal{E}(q^{(h)}||a^{(h)})\\ \nonumber
    &r = R(\mathcal{F}(G,x,h)|x) = \mathrm{Eval}(\hat{y},y),\ h \sim \pi_{\theta}(h|x) = MLP(e^{(x)},e^{(h)})
\end{align}
}
\vspace{-6mm}

\noindent where $\mathcal{E}$ is the encoding model that maps text into the embedding space, $\pi_{\theta}$ is the policy model parameterized by $\theta$, and $h$ represents an selected record consisting of the question $q^{(h)}$ and the answer $a^{(h)}$ from the Q\&A database. $\mathrm{Eval}$ is the reward evaluation function that compares the human grading results with those generated by the LLM-based grader $\mathcal{F}$ after introducing the selected question and answer pairs. The diagram of the RL-training is presented in Figure~\ref{fig:rl_diag}.

\subsubsection{Reflector}
The \textit{Reflector} proposes ways to improve the current rubric $G_{t-1}$ by reflecting on the error samples returned by the \textit{Grader}. Unlike prior work~\cite{chu2024llm}, which only reflects on rubric text and responses, we introduce computer-human Q\&A information into the process. By referencing the retrieved Q\&A pairs, the \textit{Reflector} offers an in-depth reflection on the problems in the current rubric and provides constructive suggestions for improvement. Specifically, we use a two-step instruction prompt for the LLM to achieve this, as proposed by \textit{GradeOpt}. In the first step, the LLM analyzes individual and shared failure reasons for a set of error samples. In the second step, the LLM proposes suggestions to resolve these issues. This two-step improvement process is analogous to the gradient descent algorithm used in machine learning parameter optimization~\cite{ruder2017overviewgradientdescentoptimization}. In this analogy, the rubric $G$ acts as the parameter of the \textit{Grader}, identifying error reasons is similar to computing the `gradient', and proposing improvements based on these reasons is analogous to descending along the `gradient' to optimize $G_{t-1}$. The prompt for the \textit{Reflector} agent is shown in Figure~\ref{fig:reflect_prompt}.

\input{examples/reflector_prompt}

\subsubsection{Refiner}
The role of \textit{Refiner} is to generate a new rubric, $ G_t $, based on the suggestions provided by \textit{Reflector}. Before delving into details of the refinement process, we first introduce three key components of a typical grading guideline: question stem ($ G_{qs} $), key concept ($ G_{kc} $), and scoring rubric ($ G_{sr} $). Specifically, $ G_{qs} $ contains full content of the question, $ G_{kc} $ outlines the core knowledge concepts being tested, and $ G_{sr} $ provides operational guidance for human graders on how to score responses. \textit{Refiner} focuses on optimizing $ G $ by appending new adaption rules ($ G_{ar} $), which provide detailed explanations of reflections from failed predictions and identified errors. During each iteration, the \textit{Refiner} modifies examples and illustrations within $ G_{ar} $, making edits such as adding, removing, or altering content. Other components, such as $ G_{qs} $, $ G_{kc} $, and $ G_{sr} $, remain unchanged, as these are designed by human experts and modifications could distort the intended scoring logic. The refined rubric is then expressed as $ G_t = \{ G_{qs} \| G_{kc} \| G_{sr} \| G_{ar} \} $, where $ \| $ denotes text concatenation. The prompt for \textit{Refiner} is shown in Figure~\ref{fig:refine_prompt}.

\input{examples/refiner_prompt}

\subsection{Iterative Optimization}
\vspace{-1mm}

To integrate the three components of the proposed framework, we extend the nested iterative optimization design from GradeOpt~\cite{chu2024llm} into a three-layer iteration, maintaining optimization efficiency while minimizing human effort. The overall optimization procedure is outlined in Algorithm~\ref{algo:two-fold}. Specifically, In the outer iteration (lines 3–5), the rubric and all training samples are input into the grading component. Samples identified as erroneous are forwarded to the inquiring component, which generates questions for human graders. The top-k questions, ranked by grading confidence scores, are selected and answered by human graders. After validation, the refined human-computer Q\&A data is stored as an external database for use in subsequent middle and inner iterations. In the middle iteration (lines 8–9), instead of processing the entire training dataset, batches of labeled responses are sampled and graded by the \textit{Grader}. Responses where the grading results deviate from human-labeled scores are selected for the inner iteration. In the inner iteration (lines 11–19), the \textit{Reflector} identifies erroneous grading responses and suggests rubric improvements. These suggestions are then refined by the \textit{Refiner}, generating an updated rubric for the next inner iteration. To enhance optimization efficiency and prevent local maxima GradeHITL employs a beam search strategy (lines 14–17). In each outer iteration, $K$ independent inner iterations run in parallel, each producing $L$ candidate rubrics. At the end of the outer iteration, the top $K$ rubrics with the best grading performance are selected for the next iteration. To optimize for challenging samples, we adopt a batch sampling method similar to GradeOpt. The misconfidence metric is calculated as $\psi_i = \frac{\max_{\hat{y_i} \neq y_i} \log{P_{LLM}(\hat{y_i} | G, x_i)} }{ \log{P_{LLM}(y_i | G, x_i)} }$, where $\hat{y_i}$ is the \textit{Grader}'s prediction. The most challenging examples are selected for the current batch. In the next outer iteration, half of the batch is selected by querying similar challenging responses from the training dataset, while the other half is randomly chosen.

\input{examples/HITL_algo}

%% file: examples/grade_prompt.tex
\begin{figure}[!btph]
\begin{tcolorbox}[mybox={Grader Prompt}]
% \footnotesize
\scriptsize
\textbf{Task Description:}
    In this task, you assess teachers’ knowledge of students’ mathematical thinking by grading teacher's response to a math teaching question.

    \textbf{Question Stem:} 
    $<$question stem$>$
    
    \textbf{Key Concept:} 
    $<$key concept$>$

    \textbf{Scoring Rubrics:} 
    $<$scoring rubrics$>$

    \textbf{Adaptation Rules:} 
    $<$adaptation rules$>$ \newline

    \textbf{Output format}
    
    $<$score$>$
    
    Reasoning: $<$reasoning$>$
    
    % \textbf{Output Rules}
    
    % 1. Repalce ``$<$score$>$" with only one integer from 0, 1, or 2.
    
    % 2. Replace ``$<$reasoning$>$" with your reasoning for score assignment.
    
    % 3. You must strictly obey this output format. The first token in your response must be an integer score.
    
    \textit{Let's think step by step!}
\end{tcolorbox}
\vspace{-4.5mm}
\caption{An example of prompt to LLM-based \textit{Grader}.}
\label{fig:grade_prompt}
\end{figure}

%% file: examples/question_prompt.tex
\begin{figure}[!btph]
\begin{tcolorbox}[mybox={Inquirer Prompt}]
\scriptsize
\textbf{Task Description:}    
$<$grading task description$>$

\textbf{Question Stem:} 
$<$question stem$>$

\textbf{Key Concept:} 
$<$key concept$>$

\textbf{Scoring Rubrics:} 
$<$scoring rubrics$>$

\textbf{Scoring Examples:}
$<$scoring examples$>$

\textbf{Adaptation Rules:} 
$<$adaptation rules$>$

    \textbf{Questions?}  
    While grading, do you have any questions on how to use the given instructions to grade short answers?  Your questions must be closely related to using the given instruction to grade this short answer. 
    Your questions can be specific to the instructions or short answer, and they must be raised in order to clear your confusions regarding how to correctly score short answer. You must clearly explain your thinking process on this questions and how you need assistance. \newline

    \textbf{Output Format} 
    You must format your questions as a JSON array of objects. Each object should have a `question id' (an incrementing integer starting from 0), a `question'. 
    You must ensure that each question is clear, concise, and specific to this short answer. The output should be valid JSON, easily parsable by Python code.
\end{tcolorbox}
\vspace{-4.5mm}

\caption{An exemplar of question-asking prompt.}
\label{fig:question_prompt}
\end{figure}

%% file: examples/reflector_prompt.tex
\begin{figure}[!btph]
\begin{tcolorbox}[mybox={Reflector Prompt}]
\scriptsize
\begin{center}
         CONTEXT
     \end{center}  
    
    You are ReflectorGPT, a helpful AI agent reflecting on [adaptation rules] that is used by a classifier for a grading task. Your task is to reflect and give reasons for why [adaptation rules] have gotten [failed examples] wrong.

    \begin{center}
        OBJECTIVE
    \end{center}
    
    I'm trying to write a classifier for a grading task. The prompt contains two components: 1. [question stem], [key concept] and [scoring rubrics] (given by experts and should not be modified); 2. [adaptation rules] (to modify).\newline

    \textbf{Important Steps For Devising Rules: } Read [failed examples]. For each one of the errors, carefully perform the following steps:
    
    - Step 1: Clearly explain why the classifier made the mistakes, and provide detailed analyses of why this teacher response should not be interpreted in that wrong way. 

    - Step 2: Devise or modify [adaptation rules] for each mistake to help classifier effectively avoid the mistake and classify the teacher response into the correct category (label). \newline

    \textbf{Question Stem:} 
    $<$question stem$>$
    
    \textbf{Key Concept:} 
    $<$key concept$>$

    \textbf{Scoring Rubrics:} 
    $<$scoring rubrics$>$

    \textbf{Adaptation Rules:} 
    $<$adaptation rules$>$ \newline

    But [adaptation rules] gets the following examples wrong:
    
    \textbf{Failed Examples:}
    $<$errors$>$ \newline

    Give reasons for why [adaptation rules] could have gotten the failed examples wrong. 
    \textit{Let’s think step by step!}
\end{tcolorbox}
\vspace{-4.5mm}
\caption{An example of the prompt to \textit{Reflector}.}
\label{fig:reflect_prompt}
\end{figure}

% \vspace{-1mm}

%% file: examples/refiner_prompt.tex
\begin{figure}[!btph]
\begin{tcolorbox}[mybox={Refiner Prompt}]
\scriptsize
\begin{center} 
        CONTEXT 
    \end{center}
    
    I'm trying to write a classifier for a grading task. You are RefinerGPT, a helpful AI agent capable of refining [adaptation rules] to be used by the zero-shot classifier for giving the most accurate predictions.
    
    \begin{center} 
        OBJECTIVE 
    \end{center}
    
    The [adaptation rules] must contain patterns learned from [failed examples], explaining why the predicted score is wrong comparing to the correct label. You should generate new rules based on [error feedbacks] to help the classifier effectively avoid mistakes. Your new rules must strictly abide by [scoring rubrics] and cite details to give detailed explanations. \newline
    
    \textbf{Question Stem:} 
    $<$question stem$>$
    
    \textbf{Key Concept:} 
    $<$key concept$>$

    \textbf{Scoring Rubrics:} 
    $<$scoring rubrics$>$

    \textbf{Adaptation Rules:} 
    $<$adaptation rules$>$ \newline

    But [adaptation rules] have gotten several examples wrong, with the reasons of the problems examined as follows:

    \textbf{Failed Examples:} 
    $<$errors$>$

    \textbf{Error Feedbacks:}
    $<$error feedbacks$>$ \newline

    Based on the above information, I wrote one different improved set of [adaptation rules] for instructing the classifier to learn patterns from examples and avoid errors.  [Adaptation rules] must strictly abide by [scoring rubrics] and must use patterns/details from examples to clearly explain. However, [adaptation rules] are different from [scoring rubrics] or [key concept] - [adaptation rules] details how to correctly classify based on the patterns from failed examples. \textit{Let’s think step by step!}
\end{tcolorbox}
\vspace{-4.5mm}
\caption{An example of the prompt to \textit{Refiner}.}
\label{fig:refine_prompt}
\end{figure}

%% file: examples/HITL_algo.tex
\label{app:algo}

\begin{algorithm}
  \scriptsize
  \SetAlgoLined
  \KwData{training split of Dataset $\mathcal{D}_{train}$, validation split of Dataset $\mathcal{D}_{val}$, initial guidelines $\mathcal{G}$, outer loop $N$ ,middle loop iteration number $T$, inner loop iteration number $W$, parallel inner batch number $L$, guidelines beam size $K$.}
  \KwResult{Optimized guidelines $G_{opt}$.}
  Initialize $G_{0,T,W}=\{g^{(k)}_{0,T,W}\}=\{\mathcal{G}\}$\;
  \For{$n\gets1$ \KwTo $N$}{
      $\hat{\mathcal{Y}}_{train}^{(n)} \gets$ generate grading results for $\mathcal{D}_{train}$ by \textit{Grader} with guideline $g^{(k)}_{n-1,T,W}$\;
      $\mathcal{Q}^{(n)} \gets$ generate questions by running inquiring component over erroneous sample in $\mathcal{D}_{train}$\;
      $\mathcal{H}^{(n)} \gets$ generate Q\&A Dataset by filtering through the validating on $\mathcal{D}_{val}$\;
      Initialize $G_{n,0,W}$ = $G_{n-1,T,W}$ \;
      \For{$t\gets1$ \KwTo $T$}{
        $b_{out} \gets$ sample an outer iteration batch from $\mathcal{D}_{train}$ \;
        Initialize $G_{n,t,0}$ = $G_{n,t-1,W}$ \;
        \For{$w\gets1$ \KwTo $W$}{
          
          \For{$k\gets1$ \KwTo $K$}{
            $\hat{y}_{out} \gets$ generate grading results for $b_{out}$ by \textit{Grader} with guideline $g^{(k)}_{n,t,w}$ \;
            $e_{n,t,k} \gets$ find error graded samples from $b_{out}$ caused by guideline $g^{(k)}_{n,t,w}$ \;
            \DoParallel{
              $b_{in}^{(l)} \gets$ randomly sample an inner batch from $e_{n,t,k}$ \;
              $h_{n,t,w}^{(k,l)} \gets$ retrieve from $\mathcal{H}^{(n)}$ by \textit{Retriever} \;
              $g_{n,t,w}^{(k,l)} \gets$ inputting $b_{in}^{(l)}$, $h_{n,t,w}^{(k,l)}$ and $g_{n,t,w-1}^{(k)}$ to \textit{Reflector} and \textit{Refiner}  \;
            }

          }
          $G_{n,t,w} = \{g_{n,t,w}^{(k)} \mid 1\leq k \leq K\} \gets$ select top-$K$ performed $g_{n,t,w}^{(k,l)}$  over $\mathcal{D}_{val}$ \;
        }
        % update $G_{t+1,0}=\{g_0\}$ \;
      }
  }
  \caption{GradeHITL}
  \label{algo:two-fold}
\end{algorithm}

%% file: 5experiment.tex
\section{Experiment}
% \vspace{-2mm}

We conduct experiments to evaluate the effectiveness of GradeHITL. Our experiments aim to address the following research questions: \textbf{RQ1}: Does incorporating human-in-the-loop optimization improve performance compared to existing automated prompt optimization methods? \textbf{RQ2}: Does the RL-based retriever achieve better optimization results than heuristic-based retrieval methods?

\subsection{Dataset}

\vspace{-1mm}

To answer the research questions above, we evaluate our framework using a pedagogical dataset designed to capture nuances in short-answer responses. This dataset was collected through a national research study assessing teachers' pedagogical knowledge of mathematics. Each question is accompanied by a grading rubric specifically designed to assess key pedagogical content knowledge. Compared to existing ASAG tasks~\cite{dzikovska2013semeval,mohler2011learning}, pedagogical answer grading requires a more nuanced interpretation to understand the respondent’s thought process. Given this complexity, it is valuable to explore the performance of LLMs in handling such challenging ASAG tasks. The dataset consists of three types of questions: knowledge of mathematics teaching ($C_1$), knowledge of students’ mathematical thinking ($C_2$), and knowledge of tasks ($C_3$). To ensure a comprehensive evaluation, we select two representative questions from each category~\cite{chu2024llm}. Responses are graded on a three-point scale by two expert annotators. In cases where the two graders disagree, a third expert determines the final label. The dataset includes a total of 1,376 responses, with an average of 229 valid responses per question. Detailed statistics for each question are provided in Table~3 in Appendix.

\subsection{Baselines}
\vspace{-1mm}

To evaluate the effectiveness of GradeHITL, we compare it against several representative ASAG baseline algorithms. First, we include two non-LLM methods: SBERT~\cite{reimers2019sentencebert} with logistic regression and RoBERTa~\cite{liu2019roberta} with fine-tuning. Both algorithms have been shown to be simple yet effective solutions for ASAG tasks in prior studies~\cite{condor2021automaticSA,poulton2021explaining}. Additionally, we compare three LLM-based grading models: naive prompting, APO~\cite{pryzant2023automatic}, and GradeOpt~\cite{chu2024llm}. The naive prompting approach relies on zero-shot learning and manual prompt engineering to instruct LLMs in the grading task. Both APO and GradeOpt employ automatic prompt optimization techniques that leverage LLMs' reflective capabilities, aiming to enhance performance by summarizing existing errors. However, their refinement strategies differ: APO modifies rubrics by introducing opposite meanings of detected errors, whereas GradeOpt enhances rubric descriptions by incorporating compact, example-based refinements.

\subsection{Setting}
\vspace{-1mm}

\begin{figure}[!btph]
    \centering
    \includegraphics[width=.7\linewidth]{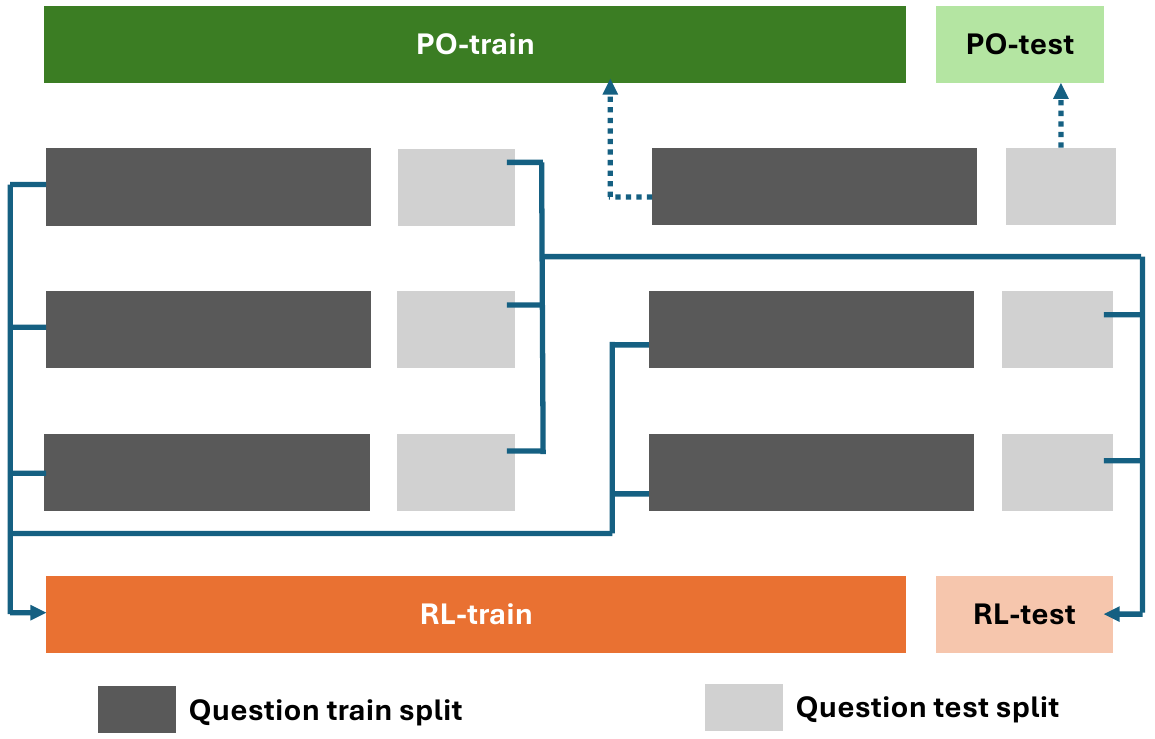}
    \vspace{-2mm}
    \caption{Data-splitting diagram for RL and prompt optimization.}
    \label{fig:data_split}
\end{figure}

To evaluate the performance of different methods, we divide each question’s responses in the dataset into training and testing sets using a 4:1 ratio. The rubric text is optimized based on the 80\% training set, while the optimized rubric is evaluated on the remaining 20\% test set. During the training of the RL-based retriever, we employ a leave-one-out strategy across the six questions. Specifically, when GradeHITL optimizes the rubric for a particular question, samples from the other five questions are used for reinforcement learning training. This approach ensures a more robust validation of the RL retriever when encountering unseen questions. The data-splitting diagram is shown in Figure~\ref{fig:data_split}. To collect human answers, we invite a grading expert to respond to the questions generated by GradeHITL. We set the outer loop to $ N=1 $, the intermediate loop to $ T=5 $ iterations, and the inner loop to $ W=3 $ iterations. A beam search selection mechanism is implemented using the Upper Confidence Bound (UCB)~\cite{auer2003using}, with a beam size of $ K=4 $. Cohen’s Kappa is used as the evaluation metric for UCB, as it has been empirically shown to outperform other metrics. 

% \vspace{-5mm}

% \vspace{-2mm}

All agents in our framework are powered by GPT-4o~\cite{hurst2024gpt} with zero-shot prompting. The \textit{Grader} operates with a temperature setting of 0.0 to minimize randomness, while both \textit{Reflector} and \textit{Refiner} use a temperature of 0.5 to encourage broader exploration of error patterns and rubric refinements. For each question, we run the algorithm three times and report the average results. Regarding retriever training, we configure the RL-based retriever with a training size of 50, a candidate size of 30, and a shot size of $ k=2 $. For nested iterative optimization, we use an outer batch size of $ |b_{out}| = 64 $ and an inner batch size of $ |b_{in}| = 8 $. For baseline methods, we follow the settings of the original studies. To ensure a fair comparison, all experiments use GPT-4o. The exploration of other LLMs is left for future work.

\subsection{Evaluation Metrics}
\vspace{-1mm}
We evaluate model performance using three metrics: Accuracy ($Acc$), Cohen's Kappa ($\kappa$), and Quadratic Weighted Kappa ($\kappa_w$). Specifically, accuracy measures the percentage of correct predictions among all predictions, $\kappa_c$ assesses the agreement between the model’s predictions and expert grading, and $\kappa_w$ further considers the differences in score values by applying a quadratic weighting function to the agreement.

\subsection{Results}
\vspace{-1mm}
To address \textbf{RQ1}, we first present the performance of GradeHITL and the baseline models in Table~\ref{tab:result_pilot}. From the results, we make the following observations. First, all four LLM-based algorithms consistently outperform the two non-LLM-based methods, highlighting the advantages of leveraging LLMs for grading tasks. Additionally, when examining metrics such as $\kappa_c$ and $\kappa_w$, we observe that representation-based methods yield \textbf{zero} values for certain questions (e.g., Q1 and Q2), despite maintaining relatively high accuracy scores. This discrepancy is typically caused by simple majority guessing in the predictions, indicating that these methods fail to learn meaningful grading patterns from the training dataset. Second, among all LLM-based algorithms, optimized prompting methods exhibit higher average performance metrics with lower variance across different questions. This suggests that rubric optimization is a crucial step in fully harnessing the potential of LLMs for automatic grading. Finally, when comparing GradeHITL to the other two fully automated rubric optimization methods, we find that GradeHITL consistently achieves the highest performance. This result demonstrates the effectiveness of incorporating HITL interactions into rubric optimization process.

\begin{table*}[!btph]
\vspace{-2mm}
\centering
\caption{Comparison of GradeHITL with baseline models. The best performed model of each metric is marked with \textbf{bold}, the second best one is marked with \underline{underline}.}
\label{tab:result_pilot}
\resizebox{\textwidth}{!}{
\begin{tabular}{@{}c|cccccc|cccccc|cccccc@{}}
\toprule
\multirow{2}{*}{\textbf{Model}} & \multicolumn{6}{c|}{\textbf{Accuracy}} & \multicolumn{6}{c|}{\textbf{Cohen's Kappa} ($\kappa_c$)} & \multicolumn{6}{c}{\textbf{Quadratic Weighted Kappa} ($\kappa_w$)} \\ \cmidrule(l){2-19} 
 & \ \ \textbf{$Q_1$} \ \ & \ \ \textbf{$Q_2$} \ \ & \ \ \textbf{$Q_3$} \ \ & \ \ \textbf{$Q_4$} \ \ & \ \ \textbf{$Q_5$} \ \ & \ \ \textbf{$Q_6$} \ \ & \ \ \textbf{$Q_1$} \ \ & \ \ \textbf{$Q_2$} \ \ & \ \ \textbf{$Q_3$} \ \ & \ \ \textbf{$Q_4$} \ \ & \ \ \textbf{$Q_5$} \ \ & \ \ \textbf{$Q_6$} \ \ & \ \ \textbf{$Q_1$} \ \ & \ \ \textbf{$Q_2$} \ \ & \ \ \textbf{$Q_3$} \ \ & \ \ \textbf{$Q_4$} \ \ & \ \ \textbf{$Q_5$} \ \ & \ \ \textbf{$Q_6$} \ \ \\ \midrule
RoBERTa & 0.76 & \textbf{0.79} & 0.45 & 0.49 & 0.55 & 0.66 & 0.00 & 0.00 & 0.00 & 0.00 & 0.00 & 0.37 & 0.00 & 0.00 & 0.00 & 0.00 & 0.00 & 0.44 \\
SBERT & 0.76 & \underline{0.74} & 0.47 & 0.76 & 0.68 & 0.62 & 0.00 & 0.00 & 0.07 & 0.17 & 0.41 & 0.29 & 0.00 & 0.00 & 0.07 & 0.17 & 0.48 & 0.33 \\ 
Naive Prompt & 0.75 & 0.51 & 0.60 & 0.70 & 0.51 & 0.66 & 0.38 & 0.09 & 0.39 & 0.55 & 0.33 & 0.48 & 0.61 & 0.24 & 0.60 & 0.62 & 0.58 & 0.68 \\
APO & 0.80 & 0.67 & 0.72 & 0.81 & 0.68 & 0.85 & 0.51 & 0.35 & 0.54 & 0.69 & 0.50 & 0.75 & 0.67 & 0.41 & 0.74 & 0.77 & 0.64 & 0.84 \\
GradeOpt & \underline{0.86} & 0.70 & \underline{0.75} & \underline{0.84} & \underline{0.73} & \underline{0.89} & \underline{0.68} & \underline{0.36} & \underline{0.56} & \underline{0.70} & \underline{0.52} & \underline{0.80} & \underline{0.76} & \underline{\textbf{0.54}} & \textbf{0.77} & \underline{0.80} & \underline{0.70} & \underline{0.87} \\ 
GradeHITL & \textbf{0.89} & 0.72 & \textbf{0.77} & \textbf{0.86} & \textbf{0.77} & \textbf{0.91} & \textbf{0.71} & \textbf{0.42} & \textbf{0.65} & \textbf{0.78} & \textbf{0.61} & \textbf{0.85} & \textbf{0.78} & \underline{\textbf{0.54}} & \underline{0.75} & \textbf{0.87} & \textbf{0.73} & \textbf{0.92} \\ \bottomrule
\end{tabular}}
\end{table*}

% \vspace{-5mm}

To address \textbf{RQ2}, we conduct ablation studies on the Q\&A selection component in GradeHITL. Specifically, we evaluate two common heuristic approaches: random selection and semantic similarity selection. The random selection method represents the most naive approach to incorporating human feedback into the reflection process, as it introduces no preference when selecting Q\&A pairs. In contrast, the semantic similarity selection method prioritizes Q\&A pairs with higher similarity to the input response. Intuitively, selecting semantically similar records should provide more relevant information to identify weaknesses in the current grading rubric, thereby improving the performance of the refined rubric. The performance comparison of these heuristic Q\&A retrieval methods and the RL-based retrieval method is presented in Table~\ref{tab:result_ablate}. From the results, we observe that the semantic similarity-based retrieval method outperforms random selection in 4 out of 6 questions, indicating that Q\&A selection significantly impacts the framework’s overall performance. Semantically relevant Q\&As are more effective in guiding the reflection process to identify rubric weaknesses. However, since semantic similarity is a heuristic method, it does not always guarantee the optimal selection. In some cases, random selection performs better, suggesting that the most semantically similar Q\&A pairs are not necessarily the most informative. Compared to the two heuristic retrieval methods, the RL-based retriever learns from collected reward values by exploring performance variations across different Q\&A pair combinations. By identifying shared patterns among high-reward pairs, it consistently selects the most valuable Q\&A pairs. As shown in Table~\ref{tab:result_ablate}, our proposed RL-based Q\&A retriever outperforms both baseline methods across all six questions, proving its effectiveness.

% \vspace{-2mm}
\begin{table*}[!btph]
\centering
\caption{Comparison of different Q\&A Retrievers: Random (RD), Semantic Similarity (SS), and Reinforcement Learning (RL). The best-performing model for each metric is highlighted in \textbf{bold}, while the second-best is indicated with \underline{underline}.}
\label{tab:result_ablate}
\resizebox{\textwidth}{!}{
\begin{tabular}{@{}c|cccccc|cccccc|cccccc@{}}
\toprule
\multirow{2}{*}{\textbf{Model}} & \multicolumn{6}{c|}{\textbf{Accuracy}} & \multicolumn{6}{c|}{\textbf{Cohen's Kappa} ($\kappa_c$)} & \multicolumn{6}{c}{\textbf{Quadratic Weighted Kappa} ($\kappa_w$)} \\ \cmidrule(l){2-19} 
 & \ \ \textbf{$Q_1$} \ \ & \ \ \textbf{$Q_2$} \ \ & \ \ \textbf{$Q_3$} \ \ & \ \ \textbf{$Q_4$} \ \ & \ \ \textbf{$Q_5$} \ \ & \ \ \textbf{$Q_6$} \ \ & \ \ \textbf{$Q_1$} \ \ & \ \ \textbf{$Q_2$} \ \ & \ \ \textbf{$Q_3$} \ \ & \ \ \textbf{$Q_4$} \ \ & \ \ \textbf{$Q_5$} \ \ & \ \ \textbf{$Q_6$} \ \ & \ \ \textbf{$Q_1$} \ \ & \ \ \textbf{$Q_2$} \ \ & \ \ \textbf{$Q_3$} \ \ & \ \ \textbf{$Q_4$} \ \ & \ \ \textbf{$Q_5$} \ \ & \ \ \textbf{$Q_6$} \ \ \\ \midrule
GradeHITL(RD) & 0.87 & \underline{\textbf{0.72}} & \underline{0.75} & 0.82 & 0.72 & 0.89 & 0.66 & \underline{0.40} & \underline{0.62} & 0.71 & 0.55 & 0.81 & 0.73 & 0.50 & \underline{0.73} & 0.81 & 0.66 & 0.87 \\
GradeHITL(SS) & \underline{\textbf{0.89}} & 0.70 & 0.73 & \underline{0.84} & \underline{0.75} & \underline{\textbf{0.91}} & \textbf{0.72} & 0.39 & 0.58 & \underline{0.74} & \underline{0.59} & \underline{0.84} & \underline{\textbf{0.78}} & \underline{0.52} & 0.72 & \underline{0.86} & \underline{0.68} & \underline{0.91} \\
GradeHITL(RL) & \underline{\textbf{0.89}} & \underline{\textbf{0.72}} & \textbf{0.77} & \textbf{0.86} & \textbf{0.77} & \underline{\textbf{0.91}} & \underline{0.71} & \textbf{0.42} & \textbf{0.65} & \textbf{0.78} & \textbf{0.61} & \textbf{0.85} & \underline{\textbf{0.78}} & \textbf{0.54} & \textbf{0.75} & \textbf{0.87} & \textbf{0.73} & \textbf{0.92} \\ \bottomrule
\end{tabular}}
\end{table*}

%% file: 6discussion.tex
\section{Discussion}
% \vspace{-2mm}
The results presented in this study show improved consistency by incorporating human feedback in automated rubric-based AI grading system. Our efforts are an exploratory stride toward creating an interactive mechanism for enhancing human alignment. It attempts to resolve the educators' challenge of drafting the optimal prompt when lack prompt engineering experience. Empirical endeavors on adapting such automated evaluation system in traditional classrooms are encouraged to fully exploit its potential. One limitation of this study is the inefficient use of the discarded less useful questions, which can be further explored to seize the values of these confusions.

%% file: 7conclusion.tex
\section{Conclusions}
% \vspace{-2mm}
In this paper, we introduce a human-in-the-loop ASAG framework that leverages LLMs’ interactive capabilities to optimize grading rubrics through expert feedback. By integrating reinforcement learning for Q\&A selection, our method effectively filters out low-quality questions, ensuring more reliable and interpretable grading outcomes. Experimental results on a benchmark dataset demonstrate that our approach outperforms existing methods, significantly improving both grading accuracy and rubric alignment. Our findings highlight the importance of incorporating human feedback in ASAG to enhance grading consistency and reliability.